\documentclass[letterpaper, 10 pt, conference]{ieeeconf}  %

\IEEEoverridecommandlockouts                              %

\usepackage[noend]{algpseudocode}
\usepackage{algorithm}
\usepackage{amsmath}
\usepackage{amssymb}
\usepackage[normalem]{ulem}
\usepackage{wrapfig}
\usepackage{bm}
\usepackage{xcolor}
\usepackage{graphicx}
\usepackage{hyperref}
\usepackage{booktabs}
\usepackage{caption}
\usepackage[frozencache]{minted}
\setminted{escapeinside=||}
\usepackage{multirow}
\usepackage{makecell}

\newcommand{\var}[1]{\textit{#1}}
\newcommand{\proc}[1]{\textsc{#1}}

\newcommand{\kw}[1]{{\bf #1}}

\newcommand{\True}[0]{\kw{True}}

\newcommand{\R}{\mathbf{R}}

\newcommand{\SE}[1]{\mathrm{SE}(#1)}

\usepackage{amsthm}
\theoremstyle{plain}
\theoremstyle{definition}
\theoremstyle{plain}
\theoremstyle{plain}

\usepackage{listings}
\definecolor{deepblue}{rgb}{0,0,0.5}
\definecolor{deepred}{rgb}{0.6,0,0}
\definecolor{magenta}{rgb}{1.0,0,1.0}
\definecolor{deepgreen}{rgb}{0,0.5,0}
\definecolor{textblue}{rgb}{.2,.2,.7}
\definecolor{textred}{rgb}{0.54,0,0}
\definecolor{textgreen}{rgb}{0,0.43,0}
\definecolor{es-blue}{rgb}{0.1372,0.666,1}
\definecolor{lightgraybg}{RGB}{247,247,247}  %

\lstdefinestyle{pythonstyle}{
    language=Python, 
    breaklines=true,
    basicstyle=\ttfamily\small,
    emphstyle=\bfseries\color{deepred}, 
    emph={forward,forward_v},         
    numbers=left,
    numberstyle=\tiny\color{gray},
    stepnumber=1,
    numbersep=12pt,
    tabsize=2,
    stringstyle=\color{textgreen},
    frame=none,                    
    columns=fullflexible,
    keepspaces=true,
    xleftmargin=\parindent,
    showstringspaces=false,
    commentstyle=\color{deepgreen},
    keywordstyle=\color{es-blue},
    backgroundcolor=\color{lightgraybg}  %
}

\newcommand{\python}[1]{\mintinline{python}{#1}}
\newcommand{\kwarg}[1]{\textcolor{deepblue}{\textit{#1}}}
\newcommand{\class}[1]{\textcolor{deepblue}{\textbf{#1}}}

\newif\ifcomments
\commentsfalse

\ifcomments
    \newcommand{\caelan}[1]{\textcolor{blue}{(CG: #1)}}
    \newcommand{\fabio}[1]{\textcolor{green}{(FR: #1)}}
    \newcommand{\todo}[1]{\textcolor{red}{(TODO: #1)}}
\else
    \newcommand{\caelan}[1]{}
    \newcommand{\fabio}[1]{}
    \newcommand{\todo}[1]{}
\fi

\title{\LARGE \bf
DemoStream: Batched Task-and-Motion-Plan Optimization \\for Automated Learning from Demonstration
}

\author{
Caelan Garrett$^{1}$, Neel Jawale$^{1}$, Fabio Ramos$^{1,2}$%
\thanks{$^{1}$NVIDIA, $^{2}$University of Sydney, {\tt\small cgarrett@nvidia.com}}%
}

\begin{document}

\maketitle
\thispagestyle{empty}
\pagestyle{empty}

\begin{abstract}
Learning from demonstration is an effective strategy for endowing robots with manipulation capabilities
but requires training on datasets comprised of massive amounts of demonstrations.
Existing demonstration collection approaches either require constant human involvement, for example, via teleoperation or are 
specialized for a single task.
Task and Motion Planning (TAMP) systems are able to autonomously solve and therefore generate demonstrations for general-purpose tasks; however, learning from their demonstrations has proven difficult due to multi-modality in their solutions.
In this work, we build on recent work leveraging GPU acceleration for TAMP and propose DemoStream, a TAMP system that optimizes demonstration quality in order to improve behavior cloning success rates.
First, we introduce batched TAMP sampling operations
along with a new TAMP algorithm that correspondingly performs batched constraint satisfaction in order to explore many solutions in parallel.
We conduct simulation experiments ablating different action spaces, cost functions, and batch sizes to analyze the effect of these choices on downstream policy success rates.
Finally, we show that we can leverage the safety guarantees of TAMP to directly collect real-world demonstrations at an average rate of 66 demos per hour. 
See our website for policy rollout videos: \href{https://demostream.github.io}{demostream.github.io}.
\end{abstract}

\section{INTRODUCTION}

Foundation models such as Vision-Language-Action Models (VLAs)~\cite{bjorck2025gr00t}
show incredible promise for robot control but require vast amounts of robot data, which is not easily obtainable. %
Teleoperation is a simple strategy for collecting robot data in the form of demonstrations but incurs a large human cost per task, 
making it difficult to scale. %
To reduce the amount of human effort, recent work developed semi-automated demonstration generation systems that adapt a handful of human-teleoperated demonstrations to new initial states~\cite{mandlekar2023mimicgen}.
However, these approaches still require annotated human demonstrations per task, preventing them from scaling autonomously to new tasks, and are typically deployed only in simulation because they produce unsafe behaviors.

Task and Motion Planning (TAMP)~\cite{Garrett2021} systems are able to autonomously solve new tasks using the same planning model.
Prior work has used TAMP to generate demonstrations for imitation learning~\cite{mcdonald2022guided,dalal2023imitating}, but due to the stochastic nature of prior TAMP algorithms, the resulting demonstrations have been difficult to learn from due to their suboptimality and multi-modality.
Recent work on TAMP leverages hardware acceleration, for example, on Graphics Processing Units (GPUs), to more efficiently solve problems~\cite{shen2025cutamp,chen2025differentiable,garrett2025schedulestream}.
We directly investigate the impact of GPU acceleration not only on producing demonstrations quickly but also on optimizing them so that they are easier to learn from.

\begin{figure}[ht!]
    \centering
    \includegraphics[trim={17cm 2cm 14cm 1cm},clip,width=0.31\linewidth]{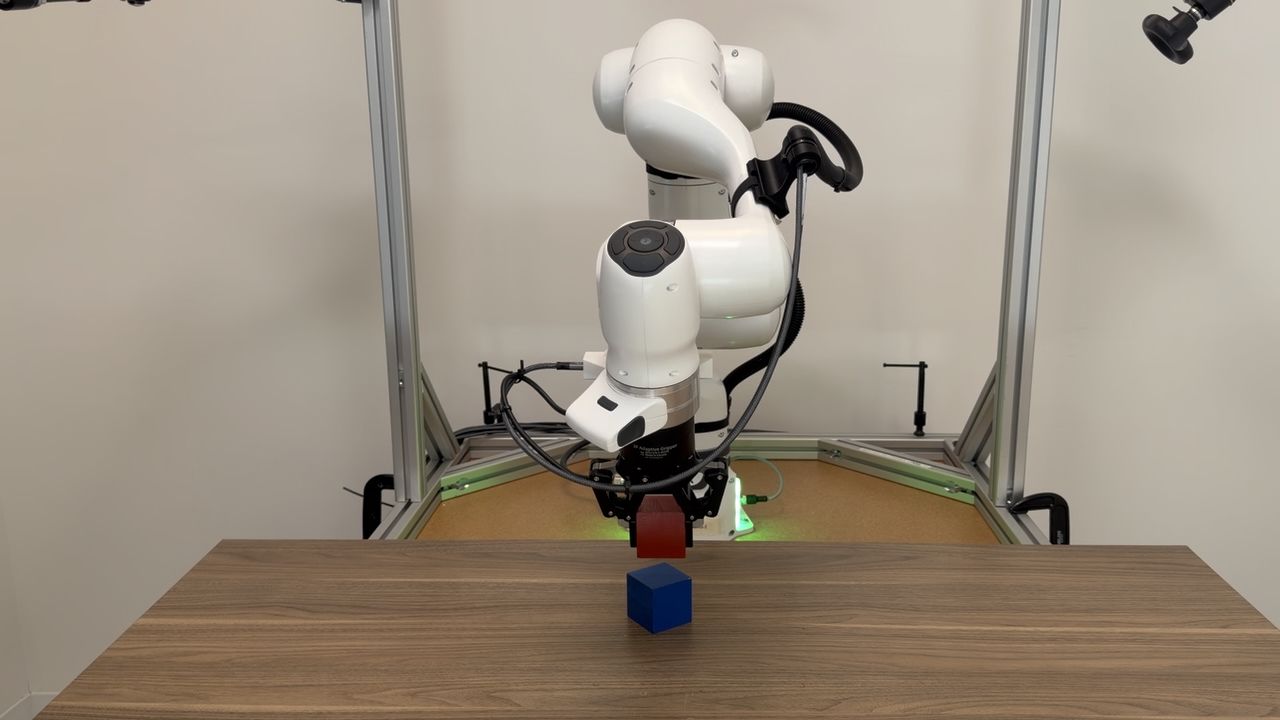}
    \includegraphics[trim={17cm 2cm 25cm 2cm},clip,width=0.37\linewidth]{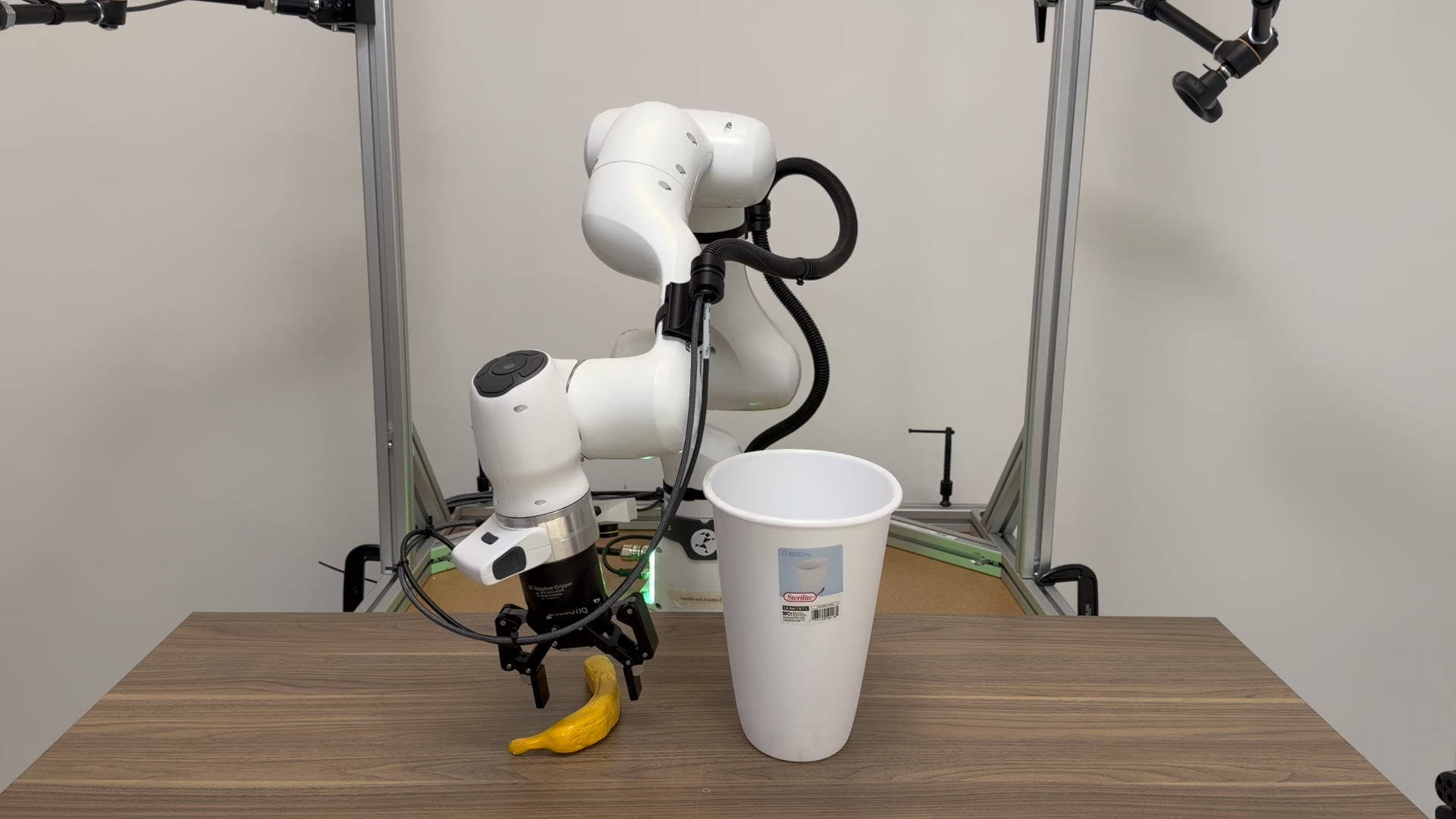}
    \includegraphics[trim={60cm 0cm 30cm 0cm},clip,width=0.295\linewidth]{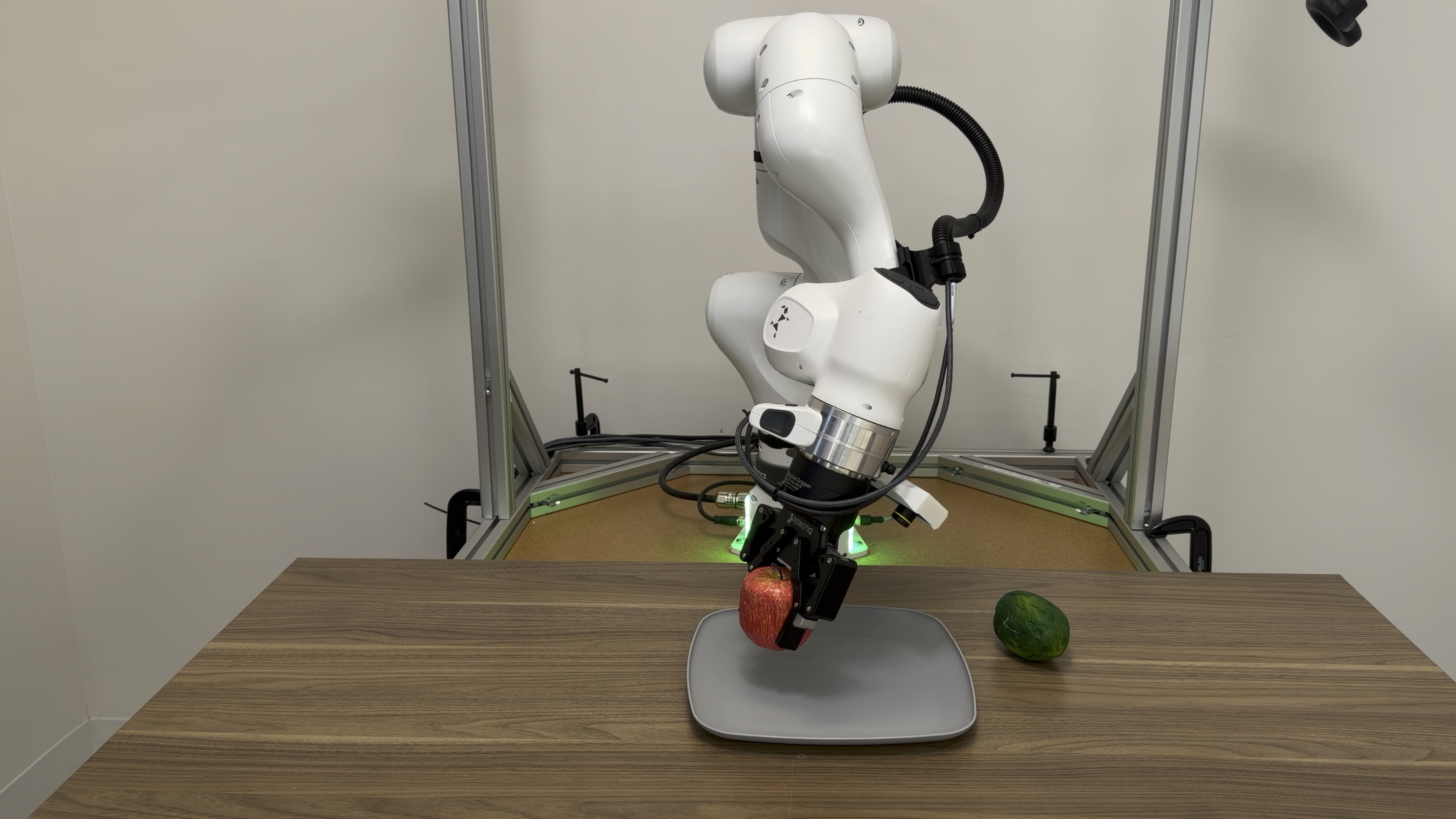}
    \caption{\small{
    \textbf{Real-World Policy Evaluation.} DemoStream collected 75 real-world demonstrations and trained visuomotor policies with a success rate of 67\%. %
    {\it Left}: stack the red block on the blue block (Stack-Two). {\it Center}: drop the banana into the trash bin (Banana-Bin). {\it Right}: pack the apple and lime on the plate (Apple-Lime).
    }}
    \vspace{-0.5cm}
    \label{fig:real}
\end{figure}

Our approach, DemoStream, first makes novel TAMP contributions in the form of {\em batched streams}, conditional samplers that generate constraint-satisfying outputs for a batch of input values, and {\em batched computation graph} optimization algorithms that compose multiple batched streams and cost functions in order to explore thousands of TAMP solutions in parallel.
These representation and algorithmic innovations give us the flexibility to explore multiple TAMP cost functions and analyze their effect on demonstration success rate, demonstration length, and downstream policy success rate.
Notably, we find that without solution optimization, TAMP systems still have high demonstration generation success rates but train policies with success rates of zero (Table~\ref{tab:actions}).
Additionally, the choice of cost function matters; we find that it is easier to learn from demonstrations that minimize task-space rather than joint-space costs, both during TAMP and motion planning, likely because these costs minimize the motion of wrist-mounted cameras, shrinking the distribution of encountered observations.
DemoStream is able to train successful visuomotor policies on a variety of tasks including multi-robot (Figure~\ref{fig:dual}), long-horizon (Figure~\ref{fig:pack}), agentic open-world (Figure~\ref{fig:arena}), and real-world (Figure~\ref{fig:real}) tasks.

\textbf{The contributions of this paper are:}
\begin{itemize}
    \item New TAMP batched samplers and computation graph optimizers for GPU-accelerated plan optimization;
    \item Automated demonstration generation via agentic, batched TAMP for end-to-end imitation learning;
    \item Analysis on the effect of action space, cost function, and batch size on policy learning success rates;
    \item Simulation experiments demonstrating long-horizon, multi-robot, and open-world DemoStream capabilities;
    \item Safe automated real-world demonstration collection.
\end{itemize}

\section{RELATED WORK}

DemoStream builds on work at the intersection of demonstration generation for imitation learning and TAMP.

\paragraph{Demonstration Generation}
We seek to acquire demonstration datasets of observation-action training examples to train visuomotor policies using Behavioral Cloning (BC)~\cite{pomerleau1989alvinn}. 
Rather than collecting these demonstrations through teleoperation~\cite{robomimic2021},
one strategy to reduce the amount of human effort involved is to teleoperate a handful of annotated demonstrations per task, adapt those demonstrations to new initial states, execute the adapted demonstrations, and retain the successful ones~\cite{mandlekar2023mimicgen}.
Non-trivial variations in the initial state induce large changes in the task-relevant workspace and safe configuration space, hampering the ability to successfully adapt demonstrations.
Several approaches interleave skill-specific demonstration adaptation with motion planning to be robust to these variations~\cite{garrett2024skillgen,yang2025physics,li2025momagen,lin2026humanoidmimicgen}.
Still, these methods are {\em task-specific} and require new human demonstrations whenever the sequence of actions or involved objects change.
In contrast, DemoStream is able to produce demonstrations for unseen tasks involving unseen objects without any human demonstrations.

\paragraph{Task and Motion Planning}
Task and Motion Planning (TAMP)~\cite{Garrett2021} jointly plans discrete \& continuous robot actions, for example, which object it should pick and how it should safely execute the pick.
Several TAMP algorithms~\cite{shen2025cutamp,garrett2025schedulestream,chen2025differentiable} leverage GPU acceleration of some kind in order to more efficiently search over continuous action parameters.
Of these, ScheduleStream~\cite{garrett2025schedulestream} is closest to our approach; however, its use of GPU acceleration focuses on efficient multi-arm trajectory collision checking.
In contrast, we introduce the notion of a batched conditional sampler, 
where a single operation generates many independent values at a time.

Two approaches~\cite{mcdonald2022guided,dalal2023imitating} investigated generating demonstrations fully autonomously using TAMP. 
Both used existing CPU-only TAMP algorithms to generate demonstrations~\cite{hadfield2016sequential,garrett2020PDDLStream} without direct attention to optimizing demonstrations for learnability.
In this work, we find that successfully solving tasks alone is not enough for successful imitation learning, one must also produce high-quality demonstrations.
Moreover, these approaches were never shown to generate real-world demonstrations in a scalable manner.

Human-In-The-Loop (HITL) TAMP~\cite{mandlekar2023hitltamp} interleaves TAMP with conventional teleoperation where TAMP delegates control to a human teleoperator during contact-rich segments of a task, such as insertion actions.
Partial autonomy enables a single human to simultaneously teleoperate multiple robot arms using a queueing system; however, a human is still present on every episode.

\section{Batched TAMP}\label{sec:tamp}

Our goal is to develop a TAMP system that can autonomously generate high-quality demonstrations for imitation learning.
To that end, we first provide background on the TAMP framework build off of (Section~\ref{sec:background}).
Then, we introduce {\em batched streams}, a new general-purpose TAMP representational capability that will allow us more efficiently optimize demonstration quality (Section~\ref{sec:streams}).
In order to maximize the utility of batched streams, we present a new TAMP algorithm that composes batched streams in {\em batched computation graphs}~(Section~\ref{sec:planning}).

\subsection{TAMP Preliminaries}~\label{sec:background}

We build our approach on top of ScheduleStream~\cite{garrett2025schedulestream}, a domain-independent language for planning \& scheduling with continuous sampling operations.
ScheduleStream extends Planning Domain Definition Language (PDDL)~\cite{mcdermott1998pddl} by offering support for Python sampling operations that generate continuous values, formalized as {\em streams}, 

To convey our approach, we will use the single-robot 
\footnote{
DemoStream also supports multiple robots (see the Multi-Arm task in Figure~\ref{fig:dual}), going beyond prior TAMP demonstration generators~\cite{mcdonald2022guided,dalal2023imitating}.}
two-object packing task in Figure~\ref{fig:pack} ({\it left}) as a running example.
In this task, the goal is to transfer both the \python{red} and \python{blue} blocks, which initially start on the \python{table}, to the \python{plate} in a manner that is efficient and avoids collisions.
In this domain, the {\em constants}, {\it i.e.}, primitive values, are the discrete names of \python{objects = [red,blue,plate,table]}, continuous robot configurations \python{q} $\in \R^d$, and continuous object placement and grasp poses \python{p}, \python{g} $\in \SE{3}$.
Discrete values are strings and continuous values are PyTorch tensors.

Relationships among constants are represented using {\em functions}. 
A function \python{|\class{Function}|} represents a relationship between a tuple of inputs, modeled as {\em parameters} \python{"?x1 ... ?xN"}, and an output.
A predicate \python{|\class{Predicate}|} is function where the output is either \python{True} or \python{False}.
Predicates and functions specify a domain {\em condition}, denoted by \python{|\kwarg{cond}|}, that conveys typing information about their parameters.
In our running example, the following {\em predicates} 
declare constant typing information about their input parameters.

\begin{footnotesize}
\begin{minted}{python}
Conf = |\class{Predicate}|("?q"); Object = |\class{Predicate}|("?obj")
Placement = |\class{Predicate}|("?obj ?p")
Grasp = |\class{Predicate}|("?obj ?g")
\end{minted}
\end{footnotesize}

\noindent
Additionally, there are {\em fluent} functions that model changing state-variable values; \python{At} is the current robot configuration, \python{Holding} is what object the robot is holding (or \python{None}), \python{Attached} gives the parent frame of object \python{"?obj"} in the kinematic tree, and \python{Pose} is $\SE{3}$ pose of object \python{"?obj"} with respect to its parent frame.

\begin{footnotesize}
\begin{minted}{python}
At = |\class{Function}|(); Holding = |\class{Function}|()
Attached = |\class{Function}|("?obj", |\kwarg{cond}|=[Object("?obj")])
Pose = |\class{Function}|("?obj", |\kwarg{cond}|=[Object("?obj")])
\end{minted}
\end{footnotesize}

States are {\em assignments} (denoted by \python{<=}) of output values to all function instances.
Any undeclared function and predicate values are implicitly \python{None} and \python{False} respectively.
The initial state in our example is:

\begin{footnotesize}
\begin{minted}{python}
initial_state = [At()<=q0, Holding()<=None,
  Pose(red)<=p_r, Pose(blue)<=p_b, Pose(plate)<=p_p, 
  Attached(red)<=table, Attached(blue)<=table, ...],
\end{minted}
\end{footnotesize}

\noindent
where \python{q0} is the initial robot configuration and \python{p_r}, \python{p_b}, \python{p_p} are the initial poses of the \python{red}, \python{blue}, and \python{plate} objects.
The goal conditions (denoted by \python{==}) are:

\begin{footnotesize}
\begin{minted}{python}
[Attached(red)==plate, Attached(blue)==plate].
\end{minted}
\end{footnotesize}

Actions describe legal decisions the system can take to change the current state.
An action \python{|\class{Action}|} has a tuple of input {\em parameters} \python{|\kwarg{params}|}, a list of equality {\em conditions} \python{|\kwarg{cond}|}, a list of assignment {\em effects} \python{|\kwarg{eff}|}, and a list of additive {\em costs} \python{|\kwarg{cost}|}.
An action can be applied to a state if all the conditions \python{|\kwarg{cond}|} are satisfied by the state.
After application, the assignments in the state are updated to have the values listed in the effects \python{|\kwarg{eff}|} and the costs in \python{|\kwarg{cost}|} are incurred.

In our example, we specify three actions [\python{move}, \python{pick}, \python{place}] although additional ones can readily be added. %
The \python{move} action has parameters for the current configuration \python{"?q1"}, trajectory \python{"?t"}, and new configuration \python{"?q2"}.

\begin{footnotesize}
\begin{minted}{python}
move = |\class{Action}|(|\kwarg{params}|="?q1 ?t ?q2",
  |\kwarg{cond}|=[Motion("?q1 ?t ?q2"), At()=="?q1", ...],
  |\kwarg{eff}|=[At()<="?q2"],
  |\kwarg{cost}|=[Distance("?q1, ?q2")])
\end{minted}
\end{footnotesize}

\noindent
The \python{pick} action has parameters for the current robot configuration \python{"?q1"}, object to be picked \python{"?obj"}, current placement pose \python{"?p"}, and new grasp pose \python{"?g"}.

\begin{footnotesize}
\begin{minted}{python}
pick = |\class{Action}|(|\kwarg{params}|="?q ?obj ?g ?p",
  |\kwarg{cond}|=[Kin("?q ?obj ?g ?p"), Holding()==None, 
        Pose("?obj")=="?p", At()=="?q"],
  |\kwarg{eff}|=[Holding()<="?obj", Pose("?obj")<="?g", 
       Attached("?obj")<=None])
\end{minted}
\end{footnotesize}

The \python{place} action has parameters for the current robot configuration \python{"?q1"}, object to be placed \python{"?obj"}, current grasp pose \python{"?g"},  new placement pose \python{"?p"}, object to be placed upon \python{"?o2"}, and the current pose of \python{"?o2"}.

\begin{footnotesize}
\begin{minted}{python}
place = |\class{Action}|(|\kwarg{params}|="?q ?obj ?g ?p ?o2 ?p2",
  |\kwarg{cond}|=[Kin("?q ?obj ?g ?p"), Holding()=="?obj", 
 Support("?obj ?p ?o2 ?p2"), Pose("?o2")=="?p2",
 Pose("?obj")=="?g", At()=="?q"] + 
 [|$\sim$|Collision("?obj", Pose("?obj"), o, Pose(o)) 
  for o in objects],
  |\kwarg{eff}|=[Holding()<=None, Pose("?obj")<="?p", 
       Attached("?obj")<="?o2"])
\end{minted}
\end{footnotesize}

\subsection{Batch Streams}~\label{sec:streams}

So far, we've modeled domains purely declaratively; however, we require a procedural component to generate values that serve as valid parameters \python{|\kwarg{params}|} for actions [\python{move}, \python{pick}, \python{place}].
Additionally, these values must satisfy the following static predicates, which are present in function conditions and action preconditions \python{|\kwarg{cond}|}.
\python{Grasp} defines stable grasps \python{"?g"}, \python{Support} defines placement poses \python{"?p"} supported by another pose \python{"?p2"}, \python{Kin} models the kinematic constraint that the end-effector pose at configuration \python{"?q"} times grasp pose \python{"?g"} gives placement pose \python{"?p"}, and \python{Motion} asserts configurations \python{"?q1"}, \python{"?q2"} are the start and end of trajectory \python{"?t"}. 

\begin{footnotesize}
\begin{minted}{python}
Grasp = |\class{Predicate}|("?obj ?g", |\kwarg{cond}|=[Object("?obj")])
Support = |\class{Predicate}|("?obj ?p ?o2 ?p2", |\kwarg{cond}|=[
  Placement("?obj ?p"), Placement("?o2 ?p2")])
Kin = |\class{Predicate}|("?q ?obj ?g ?p", |\kwarg{cond}|=[
  Conf("?q"), Grasp("?obj ?g"), Placement("?obj ?p")])
Motion = |\class{Predicate}|("?q1 ?t ?q2", |\kwarg{cond}|=[
  Conf("?q1"), Conf("?q2")])
\end{minted}
\end{footnotesize}

PDDLStream~\cite{garrett2020PDDLStream} proposed generating predicate-satisfying constants by modeling {\em streams}, conditional generators that take in a tuple of input constants and yield tuples of output constants.
A stream, denoted as \python{|\class{Stream}|}, specifies a predicate, a subset of the predicate's parameters as inputs \python{|\kwarg{cond}|} (the remaining parameters are outputs), and a Python function that returns a generator \python{|\kwarg{gen}|}.
For example, a \python{grasps} stream takes object \python{"?obj"} as input and yields grasp pose tensors for output parameter \python{"?g"}.

\begin{footnotesize}
\begin{minted}{python}
def grasp_gen(obj: str) \
    -> Iterator[Tuple[Tensor]]: ...
grasps = |\class{Stream}|(Grasp, |\kwarg{inps}|="?obj", |\kwarg{gen}|=grasp_gen)
\end{minted}
\end{footnotesize}

\begin{figure}[h!]
    \centering
    \includegraphics[trim={0cm 0cm 0cm 0cm},clip,width=1.00\linewidth]{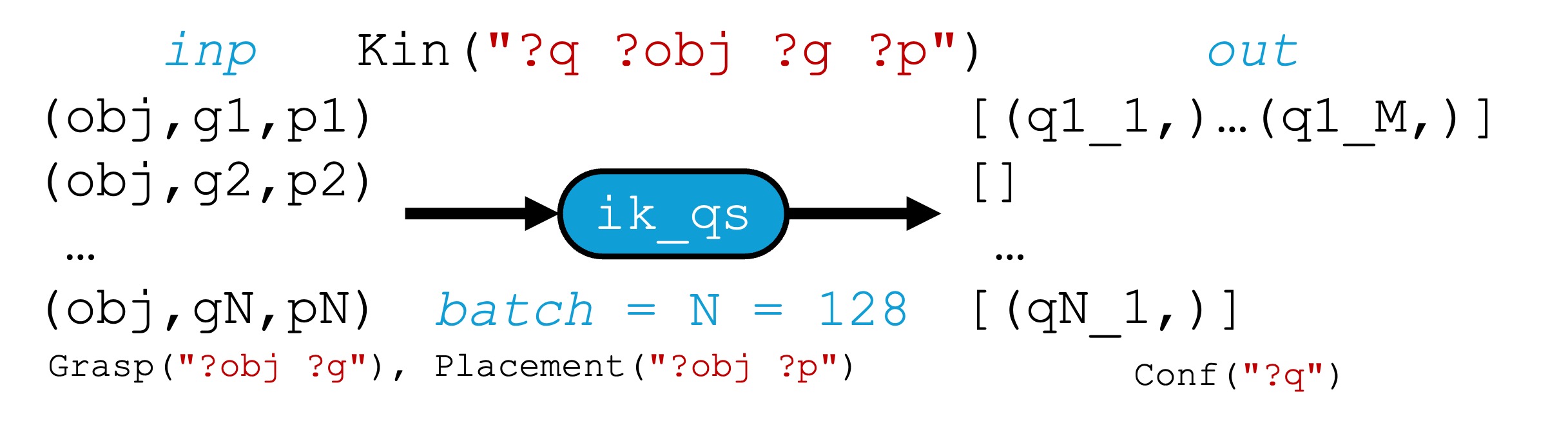}
    \caption{\small{
    \textbf{Batch IK Stream.} The \python{ik_qs} stream takes a batch of $N = 128$ inverse kinematics subproblems as inputs and returns a list of configurations \python{"?q"} per subproblem, if any can be found.
    }}
    \vspace{-10pt}
    \label{fig:ik}
\end{figure}

While it is possible to leverage GPU acceleration within \python{grasp_gen} to produce many outputs, fundamentally streams consume only a single input at a time.
This becomes limiting when we consider an Inverse Kinematics (IK) stream \python{ik_qs} responsible for satisfying \python{Kin}.
We would like to leverage advances in batched IK~\cite{curobo_icra23} to solve many IK subproblems simultaneously, for different grasp and placement combinations, but to do so, we must augment the stream contract.
To that end, we introduce {\em batch streams}.
A batch stream, \python{|\class{BatchStream}|}, specifies a batched Python function \python{|\kwarg{fn}|}, which takes in a list of input tuples and outputs a list of lists of output tuples per input tuple, along with a maximum batch size \python{|\kwarg{batch}|}.
For example, the \python{ik_qs} stream in Figure~\ref{fig:ik} takes $N = 128$ grasp and placement pairs as input and solves for configurations that satisfy them.

In our domain, we batch the placement generation, IK, and motion streams \python{placements}, \python{ik_qs}, and \python{motions} because they all take continuous values as inputs.

\begin{footnotesize}
\begin{minted}{python}
def support_fn(inp: List[Tuple[str,str,Tensor]]) \
    -> List[List[Tuple[Tensor]]]: ...
def ik_fn(inp: List[Tuple[str,Tensor,Tensor]]) \
    -> List[List[Tuple[Tensor]]]: ...
def motion_fn(inp: List[Tuple[Tensor,Tensor]]) \
    -> List[List[Tuple[Tensor]]]: ...
\end{minted}
\end{footnotesize}

\begin{footnotesize}
\begin{minted}{python}
placements = |\class{BatchStream}|(Support, 
  |\kwarg{inps}|="?obj ?o2 ?p2", |\kwarg{fn}|=support_fn, |\kwarg{batch}|=inf)
ik_qs = |\class{BatchStream}|(Kin, 
  |\kwarg{inps}|="?obj ?g ?p", |\kwarg{fn}|=ik_fn, |\kwarg{batch}|=256)
motions = |\class{BatchStream}|(Motion, 
  |\kwarg{inps}|="?q1 ?q2", |\kwarg{fn}|=motion_fn, |\kwarg{batch}|=1)
\end{minted}
\end{footnotesize}

\noindent
We also batch functions that compute configuration distances (\python{Distance}) and placement collisions (\python{Collision}).

\begin{footnotesize}
\begin{minted}{python}
def distance(inp: List[Tuple[Tensor,Tensor]]) \
    -> List[float]:
def collision(inp: List[Tuple[str,Tensor, \
    str,Tensor]]) -> List[bool]:
\end{minted}
\end{footnotesize}

\begin{footnotesize}
\begin{minted}{python}
Distance = |\class{Function}|("?q1 ?q2", 
  |\kwarg{fn}|=distance, |\kwarg{batch}|=inf)
Collision = |\class{Predicate}|("?obj1 ?p1 ?o2 ?p2", 
  |\kwarg{test}|=collision, |\kwarg{batch}|=inf)
\end{minted}
\end{footnotesize}

\subsection{Batch Planning}\label{sec:planning}

To effectively plan with the batch stream in Section~\ref{sec:batch}, we introduce a new planning algorithm \proc{batch-stream}.
It builds off of the \proc{lazy-stream} algorithm~\cite{garrett2025schedulestream} in that it plans lazily prior to calling streams to avoid unnecessary expensive operations.
Critically, it differs in how it calls streams once it has identified a plan in \proc{batch-optimization}.

\begin{algorithm}[h]
    \caption{Batch Stream Planning}
    \label{alg:batch}
    \begin{algorithmic}[1]
    \begin{footnotesize}
        \Procedure{batch-stream}{${\cal I}, {\cal G}, {\cal A}, {\cal S}; c_{\max}$}
        \State $S \gets \proc{lazy-streams}({\cal I}, {\cal S})$ \Comment{Create lazy versions of the streams}
        \While{\True}
            \State $I \gets \proc{apply-streams}({\cal I}, S)$ \Comment{Propagate lazy stream outputs}
            \State $\pi \gets \proc{search}(I, {\cal G}, {\cal A})$ \Comment{Solve a search subproblem}
            \If{$\pi = \kw{None}$}
                \State \Return $\kw{None}$ \Comment{Return failure}
            \EndIf
            \State $\psi \gets \proc{retrace-streams}({\cal I}, S, \pi, {\cal G})$ \Comment{Lazy plan parameters}
            \State $F \gets \proc{extract-costs}(\pi)$ \Comment{Extract action costs}
            \State $b_* \gets \proc{batch-optimization}(\psi, F; c_{\max})$ \Comment{Bind parameters}
            \If{$b_* \neq \kw{None}$}
                \State \Return $\proc{substitute}(\pi, b_*)$ \Comment{Return a solution}
            \EndIf
            \State $S \gets \Call{update-streams}{S, \psi}$ \Comment{Remove attempted streams}
        \EndWhile
        \EndProcedure
    \end{footnotesize}
    \end{algorithmic}
\end{algorithm}

Algorithm~\ref{alg:batch} displays the pseudocode for \proc{batch-stream}. %
It takes as input initial state ${\cal I}$, goal conditions ${\cal G}$, actions ${\cal A}$, and batched streams ${\cal S}$ as well as a maximum cost $c_{\max} \in [0, \infty)$.
First, it creates a set of {\em lazy} batched streams $S$, with placeholder output values~\cite{garrett2025schedulestream}.
On each iteration, it constructs an augmented initial state $I \supseteq {\cal I}$ using the lazy stream outputs.
Then, it solves a finite search subproblem.
If a plan $\pi$ is found, it identifies the lazy action parameters along the plan and retraces the streams associated with them, producing a {\em stream plan} $\psi \subseteq S$.
A stream plan can be thought of as a serialized {\em computation graph} for binding action parameters.
It also extracts the function costs $F$ from the plan.
Then, it calls \proc{batch-optimization}, the key subroutine that attempts to bind values to these lazy action parameters using batched stream calls (Algorithm~\ref{alg:satisfaction}).
The lowest-cost binding $b_*$ is used to bind the lazy action parameters on $\pi$, and the resulting plan is returned.
If no satisfying binding is found, the set of lazy streams $S$ is updated to remove the streams attempted in stream plan $\psi$.

\begin{algorithm}[h]
    \caption{Batch Stream Optimization}
    \label{alg:satisfaction}
    \begin{algorithmic}[1]
    \begin{footnotesize}
        \Procedure{batch-optimization}{$\psi$, $F$; $c_{\max}$}
            \State $B = [\{\;\}]$ \Comment{Initialize list of bindings with an empty binding}
            \For{$\langle s, ?x, ?y \rangle \in \psi$} \Comment{Stream $s$, input $?x$, output $?y$}
                \State $B' \gets [\;]$ \Comment{List of new bindings}
                \State $X \gets [\proc{substitute}(?x, b) \;\kw{for}\; b \in B]$ \Comment{Batch inputs}
                \For{$\langle x', y' \rangle \in \kw{zip}(X, s.\var{fn}(X))$} \Comment{Batch stream call}

                    \For{$b \in B$}
                        \If{$\proc{substitute}(?x, b) = x'$} \Comment{$x'$ in binding $b$}
                            \State $B' \gets B' + [b \cup \{?y: y'\}]$ \Comment{Augment binding}
                        \EndIf
                    \EndFor
                \EndFor
                \State $B \gets B'$ \Comment{Sort by increasing cost}
            \EndFor
            \State $C \gets [0.] * |B|$ \Comment{Initialize zero cost per binding}
            \For{$\langle f, ?z \rangle \in F$} \Comment{Function $f$, input $?z$}
                \State $Z \gets [\proc{substitute}(?z, b) \;\kw{for}\; b \in B]$ \Comment{Batch inputs}
                \State $C \gets C + f.\var{fn}(Z)$ \Comment{Batch function call}
            \EndFor
            \If{$|B| = 0 \;\kw{or}\; \kw{min}(C) > c_{\max}$}
                \State \Return $\kw{None}$ \Comment{No satisfying bindings}
            \EndIf
            \State \Return $B[\kw{argmin}(C)]$ \Comment{Return the best binding}
        \EndProcedure
    \end{footnotesize}
    \end{algorithmic}
\end{algorithm}

Algorithm~\ref{alg:satisfaction} displays the pseudocode for \proc{batch-optimization}, a algorithm for batched computation graph optimization.
Although we use it as a subroutine, it can be used for general-purpose constraint satisfaction to, for example, solve for collision-free simulation states.
It takes in a stream plan $\psi$, a set of function costs $F$, and a maximum cost $c_{\max}$. %
It maintains a list of bindings $B$, mappings from lazy action parameters to values.
It processes each stream instance $\langle s, ?x, ?y \rangle$ in the stream plan $\psi$, where $?x$ are input parameters and $?y$ are output parameters.
On each iteration, it concatenates binding values for input parameters $?x$ into an input batch $X$.
Then, it calls stream $s$, generating a batch of output values $y'$ associated with input values $x' \in X$.
A new list of bindings $B'$ is created by joining the existing bindings $B$ with the new parameter values $y'$.

After all streams in $\psi$ have been called, \proc{batch-optimization} computes the cost of each binding $C$ in batch.
In practice, this is interleaved with stream evaluation, where a function is evaluated as soon as its parameter values can be bound, so that the lowest-cost bindings are greedily prioritized in batch streams.
\proc{batch-optimization} fails if there are no bindings or if the minimum binding cost is greater than the input maximum cost $c_{\max}$.
Otherwise, it returns the minimum cost binding in $B$.
In our implementation, \proc{batch-optimization} runs on the CPU but each stream and function call runs on the GPU; however, all these operations could be performed on the GPU~\cite{chen2025differentiable} to maximize performance at the expense of flexibility.

As an example, suppose \proc{search} in \proc{batch-stream} returns the following plan $\pi$ for placing \python{red} on the \python{plate}. %

\begin{footnotesize}
\begin{minted}{python}
plan = [move(q0,"?t1","?q1"), 
  pick("?q1",red,"?g",p_r), 
  move("?q1","?t2","?q2"), 
  place("?q2",red,"?g","?p",plate,p_p)]
\end{minted}
\end{footnotesize}

\begin{figure}[h]
    \centering
    \includegraphics[trim={0cm 0cm 0cm 0cm},clip,width=0.8\linewidth]{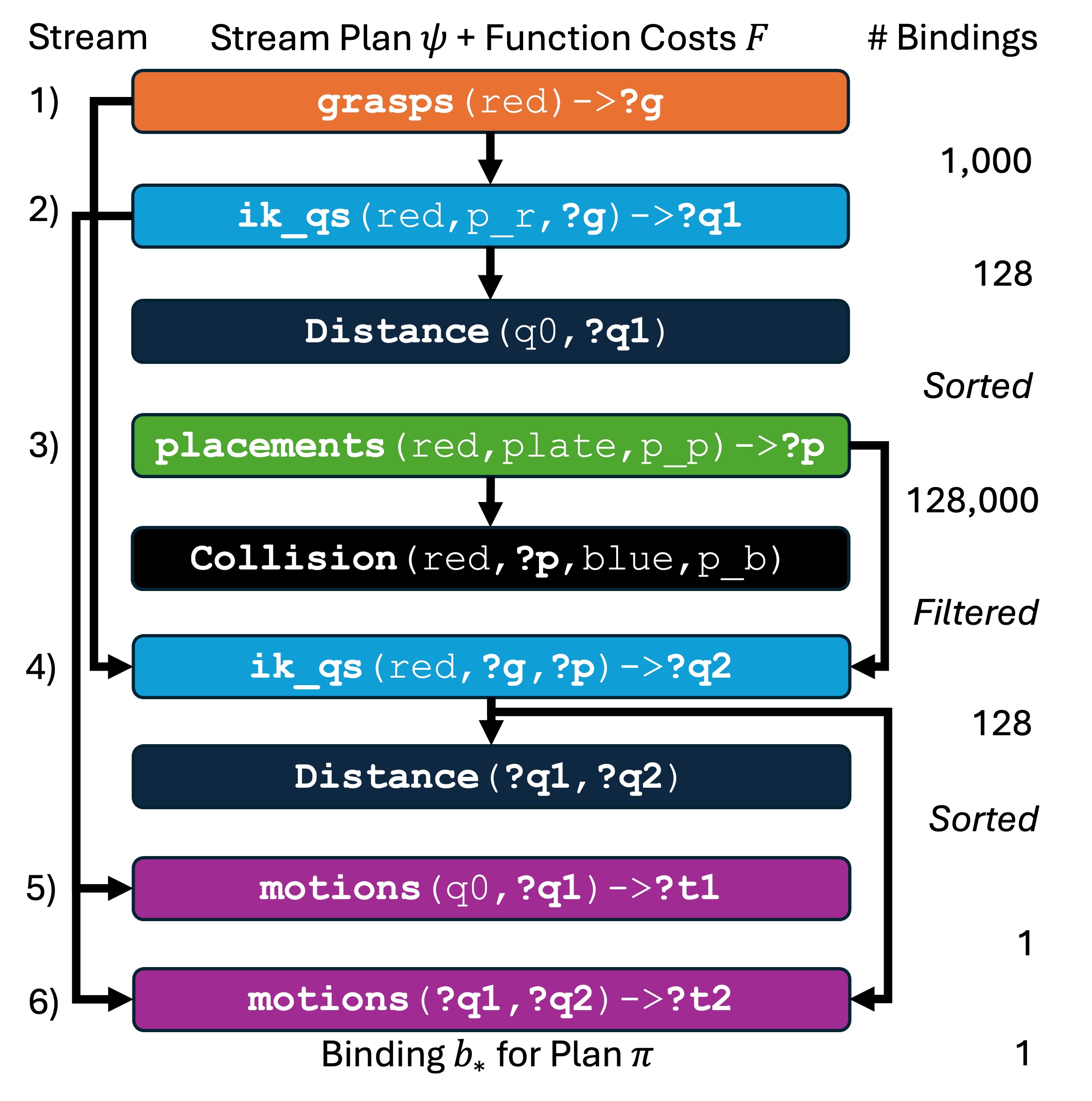}
    \caption{\small{
    \textbf{Batched Computation Graph.} The computation graph (stream plan $\psi$) corresponding to a plan $\pi$ involving picking the red block and placing it on the plate.
    The right column lists the number of bindings $|B|$ after each batch stream call.
    }}
    \vspace{-10pt}
    \label{fig:graph}
\end{figure}

\noindent
\proc{retrace-streams} and \proc{extract-costs} will produce the interleaved stream plan $\psi$ and function costs $F$ in Figure~\ref{fig:graph} corresponding to plan $\pi$.
Suppose \python{grasps} and \python{placements} each generate 1,000 outputs per input.
The bottleneck in the stream plan $\psi$ is then \python{ik_qs}, which has a batch size of 128; however, through sorting by costs and filtering by constraints, the best 128 bindings are prioritized.

\section{Demonstration Generation}

Although in Section~\ref{sec:tamp} we make novel representation and algorithmic contributions to TAMP itself, these contributions are motivated by investigating the impact of TAMP demonstration quality on visuomotor policy learning.
Here, we describe how DemoStream solves TAMP problems to produce demonstration datasets for imitation learning (Section~\ref{sec:learning}), is implemented (Section~\ref{sec:implementation}), supports multiple cost functions (Section~\ref{sec:cost}) and motion planners (Section~\ref{sec:motion}), and integrates with coding agents (Section~\ref{sec:agentic}).

\subsection{TAMP Demonstrations}\label{sec:learning}

We are interested in generating datasets of observation-action trajectories that successfully complete a task.
Here, observations consist of robot proprioception, including the joint positions, gripper position, and end-effector pose, as well as wrist and one or more external camera RGB images.
We consider three action spaces in particular to understand whether the best configuration of DemoStream depends on the action space (Section~\ref{sec:actions}).
Specifically, we experiment with joint-space absolute (Joint-Abs) and relative (Joint-Rel) actions as well as task-space relative actions (IK-Rel), implemented using an Inverse Kinematics (IK) controller.

To produce a dataset, we first interpret the goal ${\cal G}$, for example, using a Large Language Model (LLM) coding agent~\cite{kumar2024openworld}. 
Then, we sample an initial state, convert it to a planning state ${\cal I}$ and solve a TAMP problem for a plan $\pi$ that achieves the goal ${\cal G}$.
We execute plans open-loop, retain the resulting observations and actions as demonstrations, and save the demonstrations that reach a goal state
We use a fixed set of actions ${\cal A}$ and streams ${\cal S}$ across all our experiments in Section~\ref{sec:experiments}, although additional capabilities can easily be authored by LLM coding agents.
After we generate 1,000 successful demonstrations, we train discriminative~\cite{robomimic2021} and generative~\cite{chi2023diffusion} supervised models that distill the TAMP solver by mapping directly from observations to actions.

We take advantage of batched simulated environments~\cite{mittal2025isaac}
to try hundreds of TAMP solutions in parallel across different initial states.
Batched simulation negates the simulation bottleneck, leaving TAMP as the bottleneck at several seconds per successful demonstration (Figure~\ref{sec:batch}).
While we could correspondingly batch TAMP to solve multiple problems simultaneously by modeling an initial-state-generating action and stream, this comes at the expense of optimizing solution quality given a finite amount of GPU memory.

\subsection{Stream Implementation}\label{sec:implementation}

Like in prior work~\cite{curobo_icra23}, we represent the collision volume of the robot as a union of spheres greedily optimized to cover its underlying mesh model~\cite{garrett2025schedulestream}.
We use cuRoboV2~\cite{sundaralingam2026curobov2} for batched multi-pose inverse kinematics, implementing \python{ik_fn}.
We apply GPU acceleration beyond motion planning by also using it during grasp and placement generation.
We implement \python{grasp_fn} by developing a vectorized antipodal grasp sampler that generates 10,000 antipodal points on an object, samples rolls about these axes, filters out grasps that never make contact with the object, and filters out grasps in collision with the object when fully open.
We perform these collision checks efficiently by transforming just the spheres on the robot's end-effector.
Not all grasps will result in successful execution (Figure~\ref{fig:arena}); however, through trial-and-error execution, only successful grasps are retained for policy learning.
We adapt a similar strategy to implement \python{placement_fn} by randomly generating 10,000 candidate placements of object \python{"?obj"} above \python{"?o2"} using their bounding boxes, filtering out placements that do not make contact with \python{"?o2"} in the $-z$ direction, and filtering out placements that are in collision with \python{"?o2"}.
We use asymmetric collision checking between movable objects to implement \python{collision} as the conjunction of collisions between spheres of object \python{"?obj1"} and the mesh of object \python{"?o2"} and vice versa~\cite{garrett2025schedulestream}.

\subsection{Cost Functions}\label{sec:cost}

The choice of TAMP cost function to implement \python{distance}, along with how much computation is spent optimizing it, directly impacts downstream policy learning.
Optimizing a cost function implicitly enforces regularity on the distribution of TAMP solutions and thus the distribution of demonstrations.
We hypothesize that it is easier for policies to learn when the distribution of demonstrated actions per state is unimodal and concentrated, even for generative models like Diffusion Policy~\cite{chi2023diffusion}.
Concretely, the implementation of \python{distance} affects which grasps, placements, and configurations are selected within \proc{batch-optimization}, serving as targets for motion planning.

In Section~\ref{sec:actions}, we perform an experiment ablating the following three cost functions and notice dramatic effects on policy learning: {\it Zero} cost, {\it Joint}-space distance, and end-effector {\it Task}-space distance.
A zero cost function makes TAMP indifferent to the choice of solution, essentially sampling a feasible solution at random~\cite{dalal2023imitating}.
The joint-space distance is the most natural function because it correlates with the execution duration of the plan.
The end-effector task-space distance computes a weighted sum of the translation and rotation of an end-effector link.
Minimizing this cost function minimizes the motion of every link rigidly connected to the end-effector, notably wrist-mounted cameras.

\subsection{Motion Planning}\label{sec:motion}

Building on our investigation of joint- versus task-space TAMP cost functions (Section~\ref{sec:cost}), we also explored multiple ways to plan low-level motions, implementing \python{motion_fn}.
We first consider an off-the-shelf motion planner: cuRoboV2's joint-space trajectory optimizer (cuRoboV2)~\cite{sundaralingam2026curobov2}.
To have full control over planning, we also developed a batched sampling-based motion planner applicable both in joint and task space.
It takes as a hyperparameter a batched cost function modeling collision and motion edge costs.
The planner has two phases: a global exploration phase that creates a Probabilistic Roadmap (PRM)~\cite{kavraki1996probabilistic} in batch and a local exploitation phase that performs shortcutting~\cite{geraerts2007creating} in batch.
When planning in task space, we sample end-effector poses using batched forward kinematics on uniformly sampled configurations.
We check collisions only with the end-effector collision spheres, and lazily map plans to joint space using iterative IK.
We consider the following motion planners: joint-space PRM with joint-space costs (J-PRM), joint-space PRM with task-space costs (J-PRM:T), and task-space PRM with task-space costs (T-PRM).

\subsection{Agentic Goals}\label{sec:agentic} %

We are interested in generating demonstrations for tasks with novel open-world goals, for example, those described by natural language.
For example, consider the Rubiks-Near and Rubiks-Left RoboLab~\cite{yang2026robolab} tasks in Figure~\ref{fig:arena}, where the goals are for the Rubik's cube to be near and left of the bowl.
With the rise of LLM coding agents, we adopt the strategy of 
OWL-TAMP~\cite{kumar2024openworld} by parameterizing object-object goal predicates with a natural language prompt parameter.
Critically, our approach differs in that the agent must implement a batched version of the test that is compatible with efficient planning.
LLM coding agents can implement batched tests upfront using offline skills when the set of tasks is known or lazily online to respond to unseen goals. 
For simulation tasks with programmatic success tests, LLM coding agents can even read the implementation of the test and adapt them to support GPU-accelerated batched planning.

\begin{footnotesize}
\begin{minted}{python}
goal = [AgenticGoal("Rubik's cube left of the bowl", 
        rubiks, Pose(rubiks), bowl, Pose(bowl))]
\end{minted}
\end{footnotesize}

\section{EXPERIMENTS}\label{sec:experiments}

We conducted experiments to understand how the following factors affect policy Success Rates (SR): 1) action space and cost function (Section~\ref{sec:actions}), 
2) demonstration generator (Section~\ref{sec:mimicgen}),
3) batch size (Section~\ref{sec:batch}),
4) task variation (Section~\ref{sec:dual}), 5) complicated objects (Section~\ref{sec:arena}), 6) real-world constraints (Section~\ref{sec:real}).

Unless indicated otherwise, we used the IsaacLab~\cite{mittal2025isaac} physics simulator, IK-Rel action space, collected 1,000 demonstrations using DemoStream, added $[-0.01, +0.01]$ uniform noise to actions, used an IK batch size of 128, trained visuomotor BC-RNN policies~\cite{robomimic2021} or Diffusion Policies~\cite{chi2023diffusion} for 600 and 2,000 epochs,
performed 50 rollouts per policy, and reported the best-performing checkpoint from the latter half of training.

\subsection{Action Space and Cost Function}\label{sec:actions}

Our first experiment studies the effect of TAMP (Section~\ref{sec:cost}) and motion (Section~\ref{sec:motion}) cost functions on policy learning performance in different action spaces (Section~\ref{sec:learning}).
Table~\ref{tab:actions} displays the results of the experiment.
Minimizing task-space costs (Task) and using a task-space PRM (T-PRM) results in the highest success rates across action spaces.
We also trained Diffusion Policy~\cite{chi2023diffusion} on 
this data and found it had even higher success rates of 86\% (Joint-Abs), 90\% (Joint-Rel), and 86\% (IK-Rel).
The joint-space PRM with a task-space metric (J-PRM:T) outperforms cuRoboV2 and J-PRM, showing that even soft end-effector motion minimization is beneficial.
The zero cost function (Zero), which is representative of prior work~\cite{dalal2023imitating}, fails to train policies with positive success rates in both the Joint-Abs and Joint-Rel action spaces.
Additionally, joint-space costs with cuRoboV2 is representative of cuTAMP~\cite{shen2025cutamp}, which DemoStream also outperforms.
We hypothesize that minimizing wrist-camera motion is critical for effective visuomotor policy learning and note that randomized planners can generate high-quality demonstrations via batched optimization.

\begin{table}[h]
\centering
\caption{\small{\textbf{Cost Ablation.} BC-RNN policy success rates on the Stack-Three task across action spaces, cost functions, and motion planners. Minimizing task-space TAMP and motion costs maximizes policy success rate across action spaces.}}
\label{tab:actions}
\begin{tabular}{llrrrr}
\toprule
Action Space & Cost & cuRoboV2 & J-PRM & J-PRM:T & T-PRM \\
\midrule
\multirow{3}{*}{Joint-Abs} & Zero  & 0\%  & 0\%  & 0\%  & 0\% \\
& Joint & 22\% & 12\% & 32\% & 34\% \\
& Task  & 2\%  & 0\%  & 0\%  & \textbf{76\%} \\
\midrule
\multirow{3}{*}{Joint-Rel} & Zero  & 0\% & 0\%  & 0\%  & 0\% \\
& Joint & 6\% & 10\% & 26\% & 38\% \\
& Task  & 0\% & 0\%  & 0\%  & \textbf{72\%} \\
\midrule
\multirow{3}{*}{IK-Rel} & Zero  & 46\% & 20\% & 34\% & 36\% \\
& Joint & 46\% & 56\% & 76\% & 42\% \\
& Task  & 72\% & 70\% & 78\% & \textbf{82\%} \\
\bottomrule
\end{tabular}
\end{table}

\subsection{MimicGen Comparison}\label{sec:mimicgen}

To understand how DemoStream compares to existing non-TAMP demonstration generation techniques, we compared DemoStream to IsaacLab~\cite{mittal2025isaac}'s default implementation of MimicGen~\cite{mandlekar2023mimicgen} using the reference demonstrations.
Table~\ref{tab:mimicgen} shows that although each MimicGen episode is faster than a DemoStream episode, the MimicGen success rate is substantially lower than that of DemoStream (35\% vs. 92\%) and as a result the time to collect a successful demo is higher (9s vs. 6s).
Additionally, MimicGen both produced longer demonstrations because it doesn't optimize for the best feasible demonstration per problem and trained a worse-performing policy.
DemoStream outperformed MimicGen all while not requiring any annotated human demonstrations.

\begin{table}[h!]
\centering
\caption{\small{\textbf{MimicGen vs.\ DemoStream}}. DemoStream results in higher demonstration and policy Success Rates (SR) than MimicGen~\cite{mandlekar2023mimicgen}, without any task-specific annotated human demonstrations.}
\label{tab:mimicgen}
\begin{tabular}{lcccc}
\toprule
Method & \makecell{Demo SR} & \makecell{Demo Length} & \makecell{Time/Demo} & \makecell{Policy SR} \\
\midrule
MimicGen~\cite{mandlekar2023mimicgen}             & 35\% & $243\pm16$ & 9s & 34\% \\
DemoStream           & \textbf{93\%} & \textbf{$113\pm6$} & \textbf{6s} & \textbf{82\%} \\
\bottomrule
\end{tabular}
\end{table}

\subsection{Batch Size Scaling}\label{sec:batch}

\begin{figure}[h!]
    \centering
    \includegraphics[trim={0cm 0cm 0cm 0cm},clip,width=1.00\linewidth]{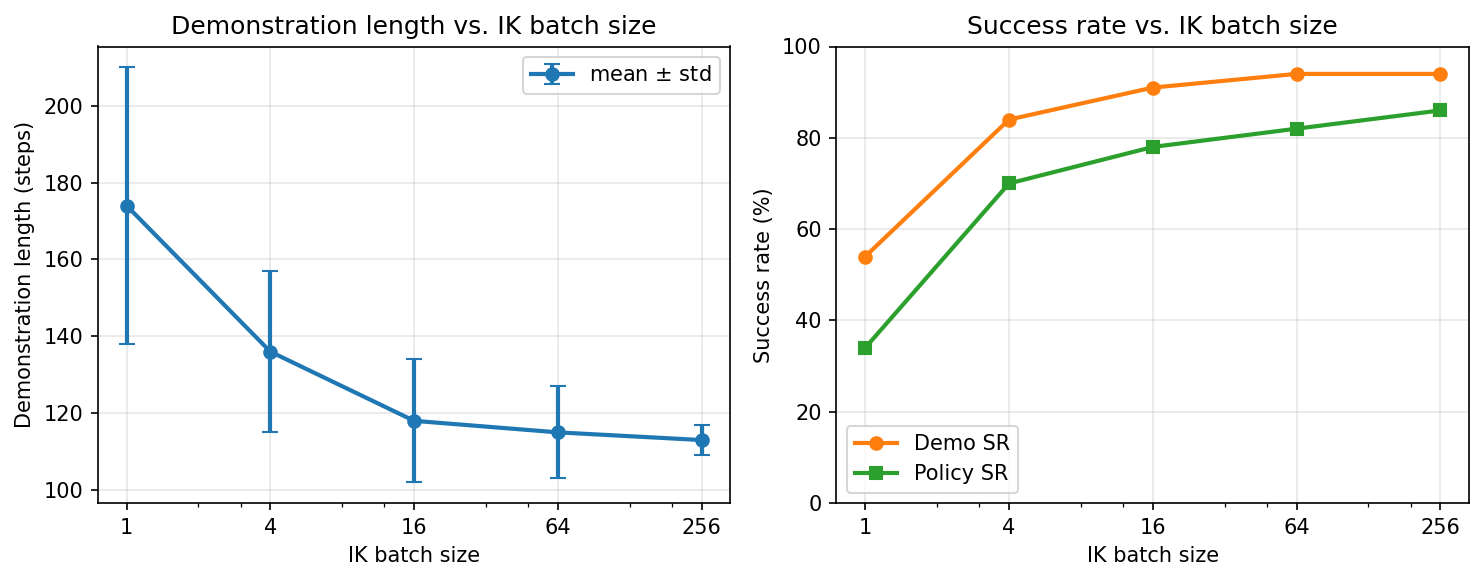}
    \caption{\small{
    \textbf{Scaling Batch Size.} The effect of scaling the IK batch size by multiples of 4. {\it Left}: the mean and standard deviation of the demonstration length. {\it Right}: the demonstration and policy Success Rate (SR). Large batch sizes decrease demonstration length and dispersion, resulting in higher policy success rates.
    }}
    \label{fig:batch}
    \vspace{-10pt}
\end{figure}

We conducted a scaling experiment to assess how DemoStream performs as the IK stream batch size is scaled from 1 to 256 by factors of 4.
Here, an IK batch size of 1 is representative of the performance of prior CPU-only TAMP methods that produce just feasible solutions~\cite{dalal2023imitating}.
Figure~\ref{fig:batch} shows that as the batch size increases, the demo success rate increases, the mean demonstration length decreases and its distribution narrows, and the policy success rate increases.
In this experiment, the mean and standard deviation of demonstration lengths correlate with policy success rate.
To understand whether simply short demonstrations were easier to learn, we halved and quartered the DemoStream execution speed, resulting in longer demonstrations, but found that the policy success rate actually slightly increased to 90\% and 92\% respectively.
Thus, mean demonstration length alone is not a good predictor of policy success rate, but Coefficient of Variation, the ratio of the standard deviation to the mean, correlates meaningfully with policy success rate.

\subsection{TAMP Tasks}\label{sec:dual} %

\begin{figure}[h!]
    \centering
    \includegraphics[trim={0cm 0cm 0cm 0cm},clip,width=0.32\linewidth]{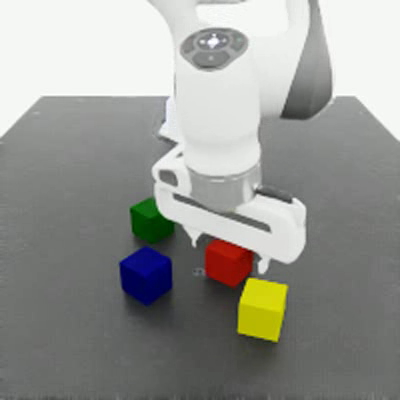}
    \includegraphics[trim={0cm 0cm 0cm 0cm},clip,width=0.32\linewidth]{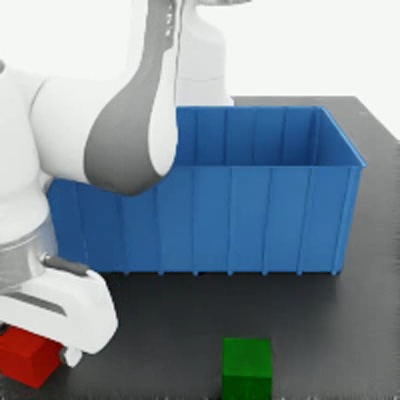}
    \includegraphics[trim={0cm 0cm 0cm 0cm},clip,width=0.32\linewidth]{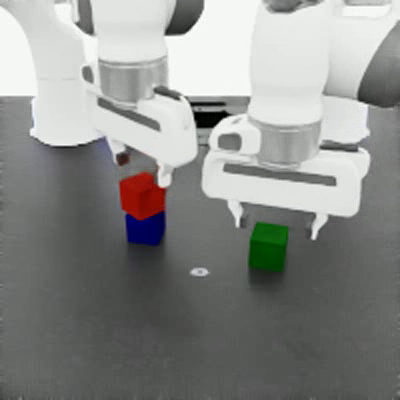}
    \caption{\small{
    \textbf{Stacking Tasks.} {\it Left}: stack four blocks (Stack-Four, long-horizon). {\it Center}: stack three blocks in the bin (Stack-Bin, obstacles). {\it Right}: stack three blocks using two arms (Multi-Arm).
    }}
    \label{fig:dual}
\end{figure}

\begin{figure}[h!]
    \centering
    \includegraphics[trim={0cm 0cm 0cm 0cm},clip,width=0.32\linewidth]{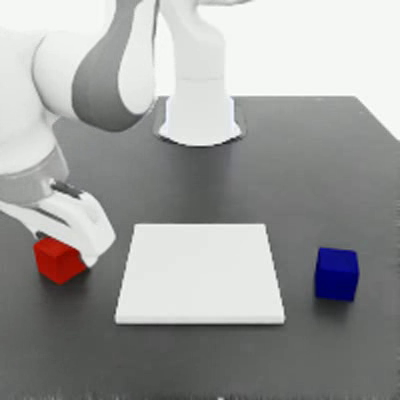}
    \includegraphics[trim={0cm 0cm 0cm 0cm},clip,width=0.32\linewidth]{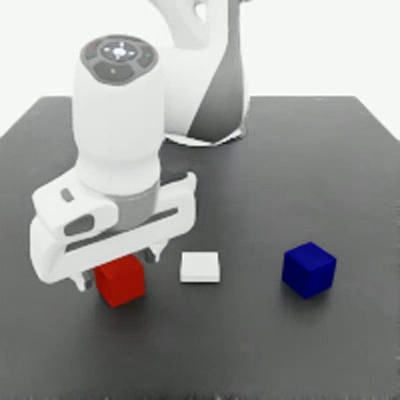}
    \includegraphics[trim={0cm 0cm 0cm 0cm},clip,width=0.32\linewidth]{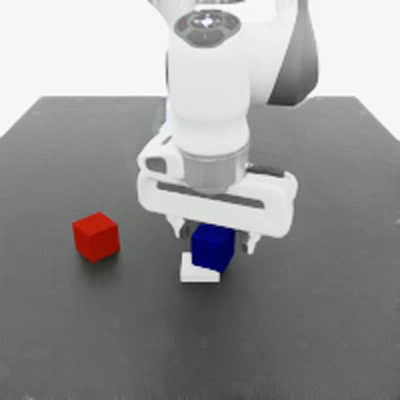}
    \caption{\small{
    \textbf{Packing Tasks.} {\it Left}: pack the two blocks on the plate (Pack-Two). {\it Center}: pack the red block on the plate (50\% Blocked). {\it Right}: pack the red block on the obstructed plate (100\% Blocked).
    }}
    \label{fig:pack}
\end{figure}

We tested DemoStream on more difficult TAMP tasks~\cite{garrett2020PDDLStream} requiring long horizons (Stack-Four), collision-aware motion (Bin-Stack), collision-aware placement (Pack-Two, 100\% Blocked), varying task plans (Pack-Either, 50\% Blocked), and multiple robots (Multi-Arm).
We defined four stacking tasks (Figure~\ref{fig:dual}): stack two (Stack-Two) and four blocks (Stack-Four), stack three blocks in a bin (Bin-Stack), stack three blocks using two arms (Multi-Arm).
We used absolute joint-space actions in Multi-Arm.
We also considered the following four packing tasks (Figure~\ref{fig:pack}): pack two blocks on the plate (Pack-Two), pack either block on the plate (Pack-Either), and pack the red block where the blue block starts on the plate with 50\% (50\% Blocked) and (100\% Blocked) probability.
Table~\ref{tab:dual} shows that DemoStream is able to train high-performing policies despite increased task difficulty.

\begin{table}[h!]
\centering
\caption{\small{\textbf{Stacking and Packing Results.} The Demonstration Success Rate (Demo SR) and policy Success Rate (SR) across the stacking (Figure~\ref{fig:dual}) and packing (Figure~\ref{fig:pack}) tasks.}}
\label{tab:dual}
\begin{tabular}{lccc}
\toprule
Task & \makecell{Demo SR} & \makecell{BC-RNN SR} & \makecell{Diffusion SR} \\
\midrule
Stack-Two      & 100\% & 100\% & 90\% \\
Stack-Four     & 72\%  &  54\% & 64\% \\
Bin-Stack   & 97\% & 82\% & 98\% \\
Multi-Arm & 54\% & 12\% & 50\% \\
Pack-Two       & 93\%  &  94\% & 96\% \\
Pack-Either    & 94\%  &  90\% & 82\% \\
50\% Blocked   & 99\%  &  88\% & 86\% \\
100\% Blocked  & 99\%  &  94\% & 90\% \\
\bottomrule
\end{tabular}
\end{table}

\subsection{RoboLab Tasks}\label{sec:arena} %

\begin{figure}[h!]
    \centering
    \includegraphics[trim={0cm 0cm 0cm 0cm},clip,width=0.32\linewidth]{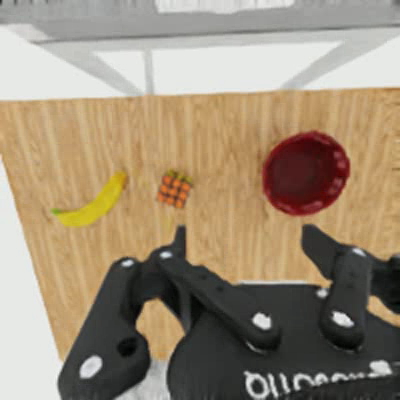}
    \includegraphics[trim={0cm 0cm 0cm 0cm},clip,width=0.32\linewidth]{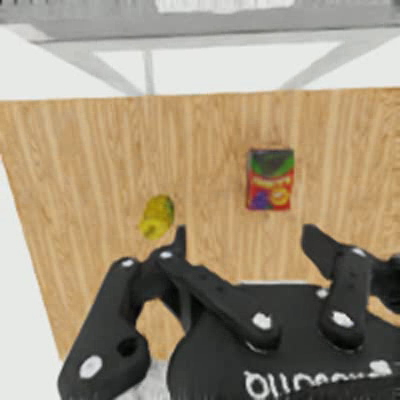}
    \includegraphics[trim={0cm 0cm 0cm 0cm},clip,width=0.32\linewidth]{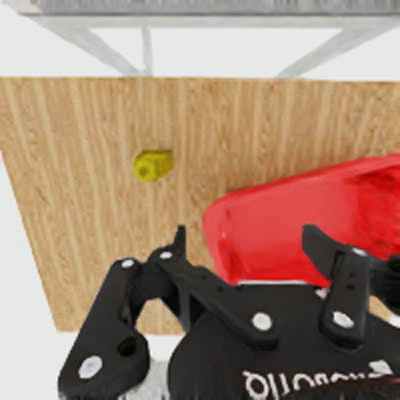}
    \caption{\small{
    \textbf{RoboLab Tasks.} {\it Left}: put the Rubik's cube and banana in the bowl (Rubiks-Banana). {\it Center}: place the mustard on the raisins (Mustard-Raisin). {\it Right}: drop the mustard in the bin (Mustard-Bin).
    }}
    \label{fig:arena}
    \vspace{-20pt}
\end{figure}

To validate that DemoStream works on more complex objects and open-world goals (Rubiks-Near, Rubiks-Left), we experimented on several RoboLab~\cite{yang2026robolab} IsaacLab Arena tasks (Figure~\ref{fig:arena}).
The tasks are: place the Rubik's cube in the bowl (Rubiks-Bowl), place the Rubik's cube near the bowl (Rubiks-Near), place the Rubik's cube left of the bowl (Rubiks-Left), place the banana in the bowl (Banana-Bowl), place the Rubik's cube and banana in the bowl (Rubiks-Banana), place the mustard bottle on the raisin box (Mustard-Raisin), and drop the mustard bottle in the bin (Mustard-Bin).
Table~\ref{tab:arena} shows that although DemoStream requires more attempts to stably grasp objects, it distills effective policies.

\begin{table}[h!]
\centering
\caption{\small{\textbf{RoboLab Results.} Demonstration Success Rate and policy Success Rate (SR) on the RoboLab tasks (Figure~\ref{fig:arena}).}}
\label{tab:arena}
\begin{tabular}{lccc}
\toprule
Task & \makecell{Demo SR} & \makecell{BC-RNN SR} & \makecell{Diffusion SR} \\
\midrule
Rubiks-Bowl           & 51\% & 86\% & 76\% \\
Rubiks-Near           & 63\% & 96\% & 56\% \\
Rubiks-Left           & 27\% & 48\% & 28\% \\
Banana-Bowl           & 40\% & 78\% & 44\% \\
Rubiks-Banana         & 3\%  & 18\% & 22\% \\
Mustard-Raisin        & 55\% & 88\% & 34\% \\
Mustard-Bin           & 17\% & 86\% & 70\% \\
\bottomrule
\end{tabular}
\end{table}

\subsection{Real World}\label{sec:real}

We conducted a real-world demonstration-collection and policy-learning experiment on a DROID~\cite{khazatsky2024droid}-style Franka Panda to understand DemoStream's viability for direct real-world learning.
The human input to the system is a command-line natural language prompt describing the task (Section~\ref{sec:agentic}).
Unlike in simulation, TAMP cannot directly read the geometry and pose of objects.
To address this, we adopted the strategy of Curtis {\it et al.}~\cite{m0m} for open-world TAMP, where object geometries are generically estimated from a point cloud segmented by SAM3~\cite{carion2026sam}.

\begin{table}[h!]
\centering
\caption{\small{\textbf{Real World Results.} The Demonstration Success Rate (Demo SR), demonstration length, time per demo, and policy Success Rate (SR) on the real-world tasks (Figure~\ref{fig:real}).}}
\label{tab:real}
\begin{tabular}{lcccc}
\toprule
Real Task & \makecell{Demo SR} & \makecell{Demo Length} & \makecell{Time/Demo} & \makecell{Policy SR} \\
\midrule
Stack-Two  & 72\% & $178 \pm 13$ & 48s & 65\% \\
Banana-Bin & 83\% & $214 \pm 19$ & 49s & 75\% \\
Apple-Lime & 79\% & $323 \pm 28$ & 69s & 60\% \\
\bottomrule
\end{tabular}
\end{table}

We considered three tasks (Figure~\ref{fig:real}): stack the red block on the blue block (Stack-Two), 
drop the banana in the bin (Banana-Bin), 
and pack the apple and lime on the plate (Apple-Lime).
DemoStream collected 75 demonstrations per task at a throughput of 66 demonstrations per hour.
The pipeline was autonomous without intervention aside from human-performed randomized resets and success labeling; for certain tasks, TAMP could automate these low-effort operations as well.
We trained a 
Joint-Abs Diffusion Policy~\cite{chi2023diffusion} 
per task.
The policy's observations were 
proprioception as well as 200x200 wrist and left-external camera images.
We evaluated the final checkpoints over 20 task trials.
Table~\ref{tab:real} shows that DemoStream trains policies that are successful on average 67\% of the time on purely real-world data.
See our website for rollout videos: \href{https://demostream.github.io}{https://demostream.github.io}.
\footnote{We disclose the use of Claude and ChatGPT to assist with code implementation, facilitating the experiments, and proofreading the manuscript.}

\addtolength{\textheight}{-12cm}   %

\bibliographystyle{IEEEtran}
\bibliography{references}

\end{document}